\documentclass[11pt]{article}

\usepackage[preprint]{acl}

\usepackage[utf8]{inputenc}
\usepackage[T1]{fontenc}
\usepackage{times}
\usepackage{latexsym}
\usepackage{hyperref}
\usepackage{url}
\usepackage{booktabs}
\usepackage{amsfonts}
\usepackage{nicefrac}
\usepackage{microtype}
\usepackage{xcolor}
\usepackage{graphicx}
\usepackage{subcaption}
\usepackage{multirow}
\usepackage{enumitem}
\usepackage{amsmath}
\usepackage{amssymb}
\usepackage{mathtools}
\usepackage{amsthm}
\usepackage[capitalize,noabbrev]{cleveref}

\theoremstyle{plain}

\theoremstyle{definition}

\theoremstyle{remark}

\title{Parallel Thinking for Large Language Model with Path Diversity}

\author{
  Jiaru Qian\textsuperscript{1,4},
  Yihong Dong\textsuperscript{2},
  Xue Jiang\textsuperscript{1,4},
  Chun-Wah Hsu\textsuperscript{1,4},
  Zhi Jin\textsuperscript{1,3},
  Ge Li\textsuperscript{1}\\
  \textsuperscript{1}School of Computer Science, Peking University \\
  \textsuperscript{2}Shanghai Jiao Tong University \\
  \textsuperscript{3}School of Computer Science, Wuhan University \\
      \textsuperscript{4}Key Laboratory of High Confidence Software Technologies, Peking University, Ministry of Education \\
  \texttt{qianjiaru26@stu.pku.edu.cn} \quad
  \texttt{lige@pku.edu.cn}
}

\begin{document}

\maketitle

\begin{abstract}
Parallel thinking has emerged as a promising paradigm for enhancing the reasoning capabilities of Large Language Models (LLMs) by exploring multiple reasoning paths simultaneously. However, we identify that parallel thinking via Reinforcement Learning (RL) typically suffers from the mode collapse of reasoning paths, where parallel thinking trajectories degenerate into similar ones. This homogeneity undermines the fundamental motivation of parallel thinking, that is, maintaining multiple hypotheses concurrently to prevent premature loss into a suboptimal solution. In this paper, we propose Diverse Parallel Thinking (DPT), a simple yet effective RL approach for parallel thinking designed to encourage path diversity. DPT rewards the model to develop distinct reasoning strategies by penalizing the prefix similarity of parallel thinking paths, thereby enhancing a more comprehensive exploration of the reasoning space.
We conduct various experiments and analyses across multiple reasoning benchmarks to demonstrate the effectiveness of DPT. Results show that DPT successfully outperforms previous parallel thinking methods, achieving state-of-the-art performance over baselines. We validate path diversity as a key element to improve parallel thinking.
Our code is  available at \url{https://github.com/JiaruQian/DPT}.
\end{abstract}

\section{Introduction}
\label{sec: intro}

Parallel thinking has emerged as a transformative paradigm for augmenting the reasoning capabilities of Large Language Models (LLMs), particularly in challenging tasks like mathematical reasoning \citep{wen2025parathinker,zheng2025parallelr1,wu2025nativeparallelreasoner,jia2025globalforktoken,tu2025deepprune}. The practical efficacy of this approach has been recently epitomized by Google's Gemini at the IMO  \citep{IMO}, where the integration of parallel and sequential thinking proved instrumental to its success~\citep{huang2025gemini}. Beyond empirical gains, parallel thinking aligns with principles in cognitive science as human experts often engage in divergent thought, maintaining multiple hypotheses concurrently to prevent premature loss into a suboptimal solution~\citep{clark1993associative, jackendoff2003precis}. By synthesizing these concurrent paths into a coherent conclusion, parallel thinking enables a more structured and deliberate reasoning process.

\begin{figure}[t]
\centering
\includegraphics[width=0.98\linewidth]{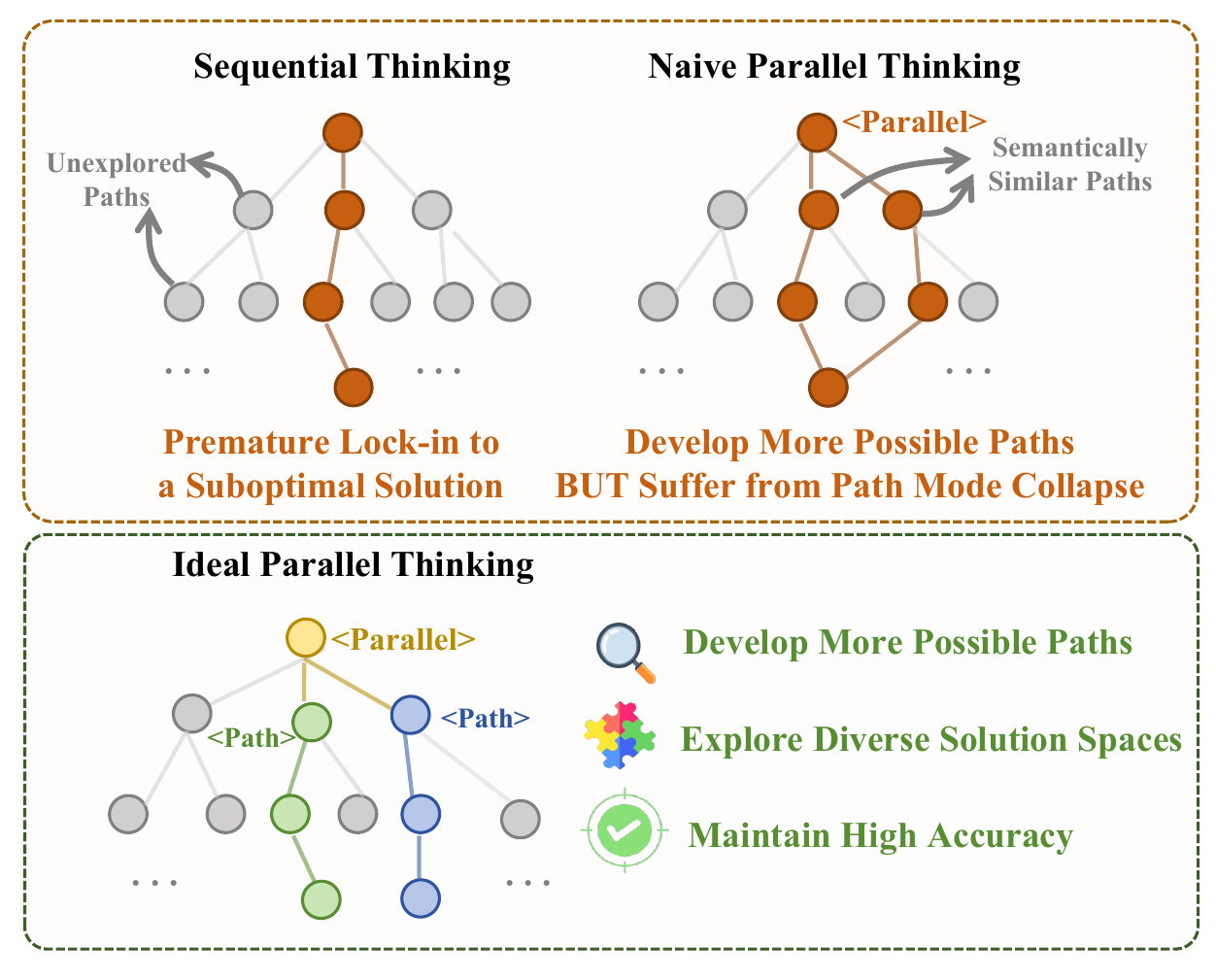}
\caption{Comparison of Traditional Sequential Thinking, Naive Parallel Thinking (RL with outcome reward), and Ideal Parallel Thinking.}
\label{fig:intro}
\vspace{-15pt}
\end{figure}

Despite this promise, we observe \textbf{mode collapse of reasoning paths} in a reproduced parallel-thinking model trained with outcome rewards. Standard RL reward formulations~\citep{yu2025rewardmodelsdeepreinforcement,gao2024designingeffectiverlreward,eschmann2021reward} optimize for correct final answers without directly rewarding diversity among internal paths. Exploitation of reliable patterns can therefore leave multiple paths highly redundant. As illustrated in Figure~\ref{fig:intro}, paths that repeatedly cover the same approach provide limited additional exploration. Our goal is to discourage this redundancy within a trajectory while preserving task correctness.

The pursuit of path diversity in parallel thinking presents two key challenges. First, the evaluation of reasoning path diversity is difficult. Second, naively maximizing diversity risks sacrificing correctness. A straightforward way is to employ lexical-level similarity like Jaccard~\citep{niwattanakul2013using} between entire paths as a metric in the training reward to encourage diversity across paths. However, such method is strongly influenced by factors such as overall path length and lexical choice, computing the reward at the whole-path level did not yield satisfactory results. Another way is to use a reward model to score the reasoning path diversity. However, it is costly and susceptible to reward hacking \citep{eisenstein2023helping,khalaf2025inference}.
Therefore, how to develop a simple yet effective metric that captures path diversity of parallel thinking remains an open problem.

In this paper, we propose Diverse Parallel Thinking (DPT), a simple yet effective RL approach for parallel thinking that enforces path diversity as an intrinsic optimization objective, thereby exploring multiple hypotheses concurrently to prevent premature convergence on suboptimal solutions. DPT integrates diversity into the reward signal through a token-level diversity metric that shifts the focus to the early part of each reasoning path. By explicitly encouraging greater diversity in the tokens generated at the beginning, we aim to steer the reasoning trajectories in different directions, allowing them to branch out and explore a broader solution space. DPT can be integrated into existing RL pipelines without requiring architectural modifications, making it applicable to parallel thinking frameworks. We evaluated DPT across a diverse range of reasoning benchmarks and various LLMs. Our results show that DPT consistently outperforms the state-of-the-art (SOTA) baseline, with ablation studies further revealing that path diversity is the primary driver behind these performance gains. These results establish path diversity as a design principle for parallel thinking in LLMs. Our main contribution can be summarized as follows:

\begin{itemize}[leftmargin=*,itemsep=2pt,topsep=2pt]
\item We identify and formally analyze the path mode collapse problem in parallel thinking, showing theoretically that outcome-supervised RL inherently drives parallel paths toward identical trajectories due to exploitation dynamics.
\item We propose DPT, a general diversity-aware RL framework for parallel thinking. DPT introduces prefix-based diversity metrics grounded in autoregressive strategy anchoring, requiring zero additional computational overhead.
\item Extensive experiments validate that DPT outperforms existing SOTA parallel thinking approaches, establishing path diversity as a key design principle for parallel thinking.
\end{itemize}

\begin{figure*}[t]
\centering
\includegraphics[width=0.95\textwidth]{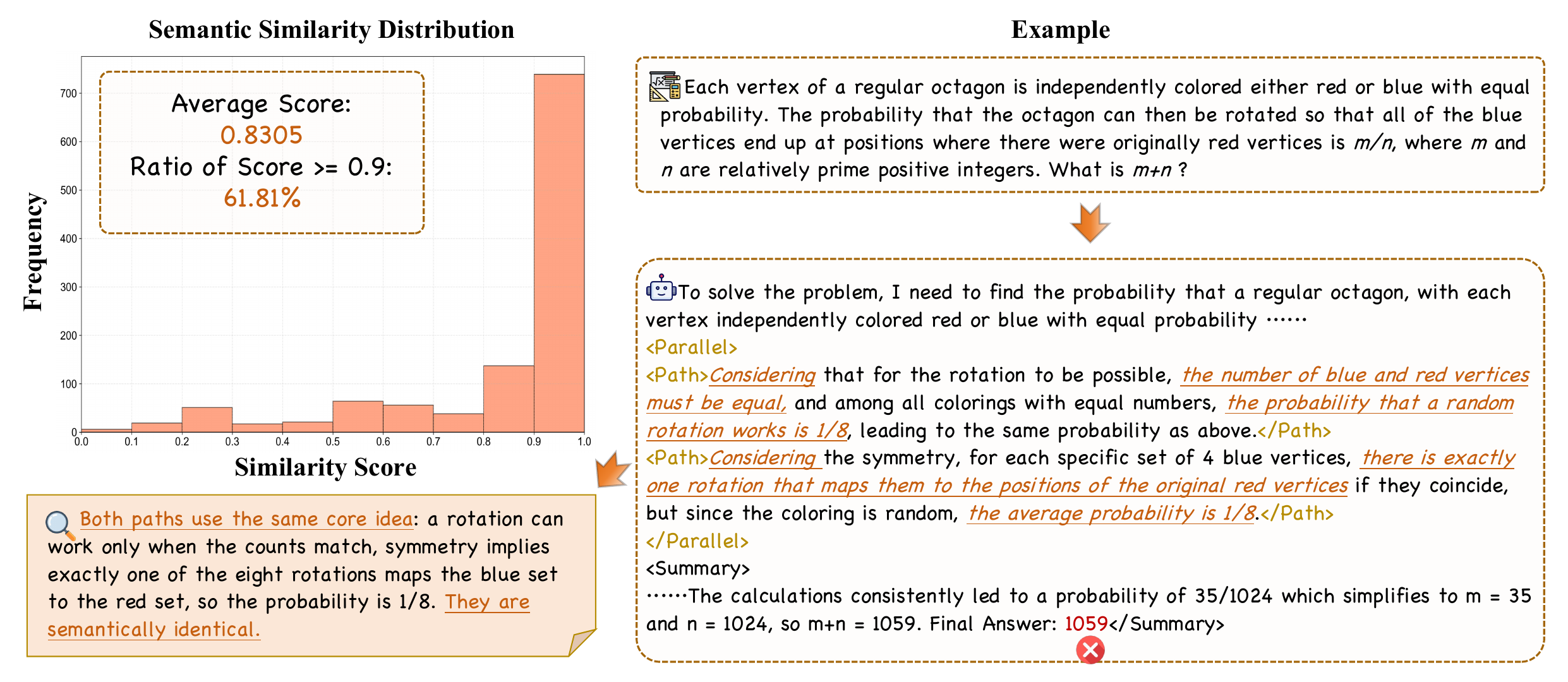}
\caption{Path Mode Collapse in Naive Parallel Thinking.}
\label{fig:motivation}
\end{figure*}

\section{Motivation}
\label{sec:motivation}
In this section, we analyze the parallel thinking models trained with the existing RL approach \citep{zheng2025parallelr1}. Despite the intuitive appeal of parallel exploration, our findings suggest that existing RL training for parallel thinking leads to the mode collapse of reasoning paths.

\noindent \textbf{Empirical Evidence.} We sampled 1,156 parallel reasoning blocks from validation rollouts of the reproduced Parallel-R1 model at training step 200 on AIME24, AIME25, AMC23, and MATH. Each included \texttt{<Parallel>} span contains $K=2$ parsed \texttt{<Path>} spans; malformed or incomplete spans are excluded. GPT-5-mini~\citep{gpt-5-mini} judges solution approach, reasoning structure, mathematical technique, expression style, and overall strategy diversity. Figure~\ref{fig:motivation} (left) shows that 61.81\% of path pairs receive similarity scores $\geq 0.9$. This is a judge-based indication of overlap whose reliability depends on the evaluation protocol; Appendix~\ref{sec:lexical_semantic_diversity} distinguishes this assessment from lexical diagnostics and states its validation limits.

The result reveals a pervasive mode collapse of parallel thinking paths in the reasoning space, which converge onto identical semantic manifolds. Figure~\ref{fig:motivation} (right) provides an example of this phenomenon in a math reasoning task. Both paths anchor on the same insight, i.e., concluding that vertex alignment constrains the probability to $1/8$. Although the phrasing fluctuates, the underlying logical heuristic remains monolithic. The model essentially learns to produce high-reward paraphrases rather than exploring the disjoint solution spaces that parallel thinking is intended to cover.

\noindent \textbf{Theoretical Analysis.}
Why does standard RL drive parallel reasoning paths toward mode collapse, as observed in Figure \ref{fig:motivation}? The failure arises from the exploitation dynamics inherent in maximizing the expected reward under outcome-supervision.
Consider a generator $\pi_\theta$ that produces an ordered tuple of $K$ latent reasoning paths $\mathcal{T} = (\tau_1, \dots, \tau_K)$ given an input $x$. A deterministic aggregation function $\hat{y}(\mathcal{T})$ extracts the final answer from the summary of all paths. The standard RL objective maximizes:
\begin{equation}
\mathcal{J}(\theta) = \mathbb{E}_{x \sim \mathcal{D}} \left[ \mathbb{E}_{\mathcal{T} \sim \pi_\theta(\cdot|x)} \left[ r\!\left(\hat{y}(\mathcal{T}),\; y^*\right) \right] \right],
\label{Eq1}
\end{equation}
\begin{equation}
\begin{aligned}
    \nabla_\theta \mathcal{J} &\approx \mathbb{E}_{x \sim \mathcal{D}} \, \mathbb{E}_{\mathcal{T}} [ \nabla_\theta \log \pi_\theta(\mathcal{T}|x) \\
    &\quad \cdot \underbrace{r(\hat{y}(\mathcal{T}), y^*)}_{\text{Reward Signal}} ].
\end{aligned}
\end{equation}
where $y^*$ is the ground truth and $r(\cdot,\cdot)$ is a binary correctness metric. Crucially, the reward depends on $\pi_\theta$ only through $\hat{y}(\mathcal{T})$ and is agnostic to the diversity of $\mathcal{T}$.

The core problem stems from the \textit{exploitation dynamics} inherent in maximizing Eq. \eqref{Eq1}. Let $\tau^*$ denote the ``safe mode'', i.e., the single reasoning pattern with the highest marginal probability of yielding the correct answer under the current policy. Because all $K$ paths are sampled from the \textit{same} shared policy $\pi_\theta$, the gradient reinforces whichever pattern most reliably produces the correct answer. This drives the marginal distribution $\pi_\theta(\tau|x)$ to concentrate on $\tau^*$, causing all path slots to simultaneously collapse:
\begin{equation}
    \pi_\theta(\tau_k | x) \to \mathbb{I}[\tau_k = \tau^*], \quad \forall k \in \{1, \ldots, K\}.
\end{equation}

Since all generated paths share the same policy, the model is not incentivized to diversify and instead converges toward $\tau^*$.


\noindent \textbf{Analysis Result.} Our analysis reveals a critical design principle: Path diversity must be an explicit optimization constraint rather than an emergent byproduct. To unlock the true potential of parallel thinking to align with its motivation, we must penalize redundancy in the latent reasoning space, specifically at the prefix level, where early strategic decisions are made.

\begin{figure*}[t]
\centering
\includegraphics[width=\textwidth]{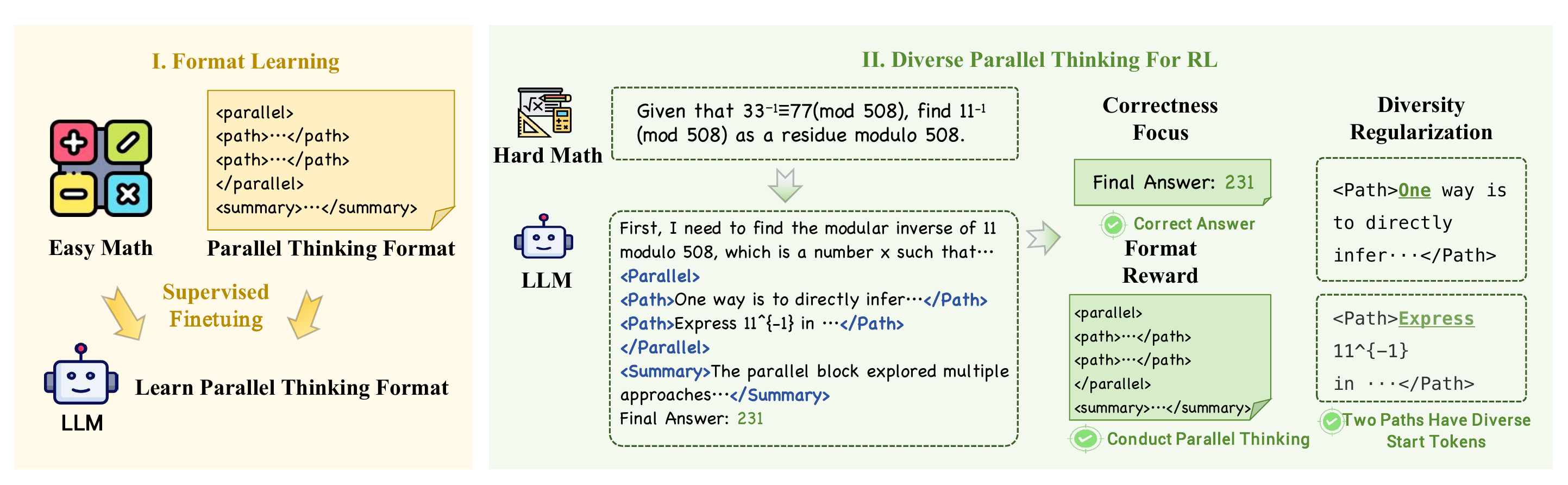}
\caption{\textbf{Overview of DPT.} (I) Format learning teaches the parallel thinking format via SFT on easy problems. (II) Diverse parallel thinking develops autonomous parallel thinking on hard problems with combined rewards.}
\label{fig:method}
\end{figure*}

\section{Related Work}
\noindent{\textbf{Enhancing LLM Reasoning Capability via RL.}} 
RL has become a central technique for enhancing LLM reasoning, with advances in policy optimization~\citep{shao2024grpo,zheng2025gspo,yu2025dapo}, on/off-policy learning~\citep{dong2025rlplus}, self-improvement without explicit labels~\citep{zuo2025ttrl}, and search-augmented reasoning~\citep{jin2025searchr1}. LTE~\citep{tang2026LTE} further explores trial-and-error learning to overcome exploration stagnation. More recently, RL with rollout diversity has gained attention. DivPO~\citep{lanchantin2025divpo}, unlikeliness reward~\citep{he2025rewardingunlikelyliftinggrpo}, and EVOL-RL~\citep{zhou2025evolrl} encourage rare, high-quality generations through reward shaping or preference optimization, while PODS~\citep{xu2025PODS}, DARLING~\citep{li2025DARLING}, and DQO~\citep{chen2025DQO} improve rollout selection and aggregation. However, these methods still treat each rollout as a sequential generation process, motivating RL for parallel inference.

\noindent{\textbf{Parallel Thinking.}}
Parallel thinking has recently emerged as a promising area in LLM reasoning. A set of works~\citep{brown2024largelanguagemonkeysscaling, huang2025selfcalibration} explores test-time compute scaling by generating multiple reasoning samples per input and aggregating them to improve answer quality. However, they typically rely on post-hoc sampling as well as simple aggregation, and tend to generate redundant or low-value traces. Recent works~\citep{jia2025globalforktoken,tu2025deepprune,wen2025parathinker,wu2025nativeparallelreasoner,zheng2025parallelr1} aim to endow models with native parallel reasoning capabilities, producing and integrating multiple thought trajectories within a single inference process.  Among them, SFT-based methods such as SSFT~\citep{jia2025globalforktoken} and ParaThinker~\citep{wen2025parathinker} inevitably lead to superficial parallel thinking, where models merely imitate parallel patterns without genuine parallel reasoning. RL-based methods such as NPR~\citep{wu2025nativeparallelreasoner} and Parallel-R1~\citep{zheng2025parallelr1} enable models to autonomously explore parallel reasoning, but often suffer from path mode collapse. 
To this end, we propose DPT to enforce path diversity while maintaining high reasoning accuracy in parallel thinking.

\section{Diversity Parallel Thinking}

We present DPT, an RL approach that enforces path diversity in parallel thinking while maintaining reasoning accuracy. We first formalize parallel thinking as a structured generation process that produces multiple independent reasoning paths within a single trajectory. We then introduce DPT's training algorithm, which integrates cold-start initialization with SFT and employs a diversity-aware RL reward function to explicitly penalize mode collapse among reasoning paths. An overview of the DPT framework is shown in Figure~\ref{fig:method}.

\subsection{Parallel Thinking Formulation}

Given a reasoning problem $x$, parallel thinking generates $K$ distinct reasoning paths $\{\tau_1, \tau_2, \ldots, \tau_K\}$ within a single coherent generation trajectory. Unlike traditional beam search or sampling that produces independent sequences, parallel thinking enables the model to adaptively decide when to branch into multiple exploration threads during the reasoning process.

We implement this behavior through three special control tags that structure the generation:
\begin{itemize}
    \item \texttt{<Parallel>...</Parallel>}: Marks the parallel exploration phase, indicating where the model suspends the main reasoning chain to explore multiple solution strategies concurrently.
    \item \texttt{<Path>...</Path>}: Delimits an internal reasoning thread intended to explore a solution approach. Information isolation depends on the attention mode.
    \item \texttt{<Summary>...</Summary>}: Synthesizes insights across parallel paths and resolves conflicts before resuming the main reasoning chain.
\end{itemize}

Formally, let $z$ denote the complete generation trajectory for problem $x$. The model autoregressively generates tokens according to policy $\pi_\theta(\cdot | x)$. When the model emits a \texttt{<Parallel>} token, it spawns $K$ parallel paths $\{\tau_1, \ldots, \tau_K\}$ within separate \texttt{<Path>} blocks. Each path $\tau_i$ is intended to explore a distinct reasoning strategy. In the ``Unseen'' mode, structural attention masking enforces true information isolation between paths; in the ``Seen'' mode, paths may attend to prior paths. A \texttt{<Summary>} block then integrates the parallel explorations before resuming the main trajectory.

The objective of parallel thinking is twofold: exploring multiple strategies increases the probability of finding correct answers, and diverse paths enable multi-perspective verification. However, standard RL optimizes only for correctness without penalizing redundancy, causing paths to collapse into identical strategies. DPT addresses this by explicitly rewarding diversity among parallel paths.

\subsection{Training Algorithm}

We train DPT through a two-stage curriculum that addresses the cold-start problem in parallel thinking. The model first learns the basic format and structure of parallel reasoning on simpler problems with supervised finetuning, then transitions to reinforcement learning on challenging tasks to enhance performance with path diversity.

\subsubsection{Cold Start Initialization}

Since pretrained language models have not encountered the special control tags (\texttt{<Parallel>}, \texttt{<Path>}, \texttt{<Summary>}) during pretraining, they cannot generate valid parallel thinking trajectories for RL exploration. We address this cold-start problem through supervised fine-tuning on a curated dataset of parallel thinking examples constructed from the GSM8K training set. We prompt a capable language model to generate solutions in the parallel thinking format and filter outputs to retain only those with valid tag structure. This phase instills familiarity with the parallel thinking format---when to trigger \texttt{<Parallel>} blocks, how to structure independent \texttt{<Path>} segments, and how to synthesize conclusions in \texttt{<Summary>} blocks---enabling subsequent RL exploration on harder tasks.

\subsubsection{Reinforcement Learning}

After cold-start initialization, we train the model using Group Relative Policy Optimization (GRPO)~\citep{shao2024grpo}, a well-known policy gradient method designed for grouped samples.

\noindent \textbf{Training Objective.} For each problem $x$, GRPO samples $N$ trajectories $\{z_1, \ldots, z_N\}$ from the policy $\pi_\theta(\cdot | x)$ and computes rewards for each trajectory. The policy gradient estimates the advantage function using group-relative normalization:
\begin{equation}
\hat{A}(z_i) = \frac{R(z_i) - \bar{R}}{\sigma_R + \epsilon},
\end{equation}
where $\bar{R} = \frac{1}{N}\sum_{j=1}^N R(z_j)$ and $\sigma_R$ are the mean and standard deviation of rewards within the group, and $\epsilon$ is a small constant for numerical stability. This normalization reduces variance compared to absolute rewards.
The policy is updated via:
\begin{equation}
\mathcal{L}(\theta) = -\mathbb{E}_{x \sim \mathcal{D}, z_i \sim \pi_\theta(\cdot|x)} \left[ \hat{A}(z_i) \log \pi_\theta(z_i | x) \right],
\end{equation}
subject to a KL divergence constraint $\text{KL}(\pi_\theta || \pi_{\text{ref}}) < \delta$ to prevent the policy from deviating too far from the reference model $\pi_{\text{ref}}$. We use the pretrained base model as the reference.

\noindent \textbf{Reward Modeling.} The reward function $R(z_i)$ determines what behaviors the model learns. Here $z_i$ denotes the complete generation trajectory, which may contain multiple parallel paths $\{\tau_1, \ldots, \tau_K\}$. Standard RL for reasoning uses only a correctness reward $r_c(z_i) = \mathbb{I}[\hat{y}(z_i) = y^*]$, where $\hat{y}(z_i)$ is the final answer extracted from trajectory $z_i$ and $y^*$ is the ground truth. Existing parallel thinking approaches also introduce format reward \(r_f(z_i)\):
\[
\begin{aligned}
 r_f(z_i) = \mathbb{I}[&\text{At least one correct}\\
 &\text{\texttt{<Parallel>} block in } z_i].
\end{aligned}
\]
However, optimizing correctness alone causes mode collapse. The model learns to exploit a single high-reward strategy, generating semantically identical paths despite the parallel thinking structure.

DPT addresses this by incorporating a diversity reward that explicitly penalizes mode collapse. For each trajectory $z_i$, the total reward combines correctness and diversity of all parallel paths \(\{\tau_1, \tau_2, \ldots, \tau_K\}\).
\begin{equation}
R(z_i) = r_c(z_i) + \lambda_f \cdot r_f
(z_i)+ \lambda_d \cdot r_d(\{\tau_1, \ldots, \tau_K\}),
\end{equation}
where $\lambda_d$ controls the trade-off between correctness and diversity, and $r_d$ measures the diversity across all $K$ parallel paths generated for problem $x$.

\noindent \textbf{Diversity Metric Design.} Measuring trajectory diversity presents a key challenge. When parallel paths share the same preceding context, the model tends to fall into similar reasoning patterns despite independent generation processes. A natural approach is to compute similarity metrics over entire paths, such as Jaccard similarity between complete token sequences. However, whole-path metrics suffer from two fundamental issues. First, path length creates spurious diversity signals. Two paths employing identical reasoning strategies but differing in verbosity yield low similarity scores, falsely indicating strategic divergence. Second, lexical choice confounds semantic similarity. Paths using the same underlying approach but varying word choices appear diverse under token-level metrics, rewarding superficial paraphrasing rather than strategic exploration.

We therefore shift focus to the early part of each reasoning path. In autoregressive generation, each prefix conditions subsequent tokens through $P(x_t | x_{<t})$. Encouraging different prefixes supplies a local symmetry-breaking signal that can create opportunities for different continuations. However, paths with different openings can still use the same strategy, and paths with shared openings can diverge later. Prefix divergence is therefore an inexpensive training proxy rather than a sufficient condition for semantic strategy diversity. We compare the first $n$ content tokens of each path pair:
\begin{equation}
d(\tau_i, \tau_j) = \mathbb{I}[\text{prefix}(\tau_i, n) \neq \text{prefix}(\tau_j, n)],
\end{equation}
where $\text{prefix}(\tau, n)$ extracts the first $n$ content tokens from path $\tau$ (starting after the \texttt{<Path>} tag). This binary metric returns 1 if the prefixes differ and 0 if they are identical.
We also explore a similarity variant that measures token-level overlap:
$d_{\text{sim}}(\tau_i, \tau_j) = 1 - \frac{|\{t : \text{prefix}(\tau_i, n)_t = \text{prefix}(\tau_j, n)_t\}|}{n}$,
which provides a finer-grained signal. We ablate both variants in Section~\ref{sec: full_ablation}.

The diversity reward aggregates pairwise comparisons across all $K$ paths ($K \geq 2$):
\begin{equation}
r_d(\{\tau_1, \ldots, \tau_K\}) = \frac{1}{\binom{K}{2}} \sum_{i=1}^{K-1} \sum_{j=i+1}^{K} d(\tau_i, \tau_j).
\end{equation}
This normalized score ranges from 0 (all paths identical) to 1 (all paths distinct). The metric penalizes mode collapse by rewarding models that generate diverse prefixes. Token-level comparison offers several advantages. Computational efficiency improves because we do not need to encode full trajectories into dense representations. Interpretability improves because differences are human-verifiable at the token level. The early divergence focus captures strategic differences effectively, as high-level approach selection manifests in the initial tokens.

\paragraph{Training Curriculum.} If diversity pressure is applied too aggressively, the model may prioritize superficially different paths over correct answers. During RL training, the full reward $R(z_i) = r_c(z_i) + \lambda_f \cdot r_f(z_i) + \lambda_d \cdot r_d(\{\tau_1, \ldots, \tau_K\})$ is applied only for the last 2 steps of every 10 training steps, while the remaining 8 steps use the correctness reward $r_c(z_i)$ alone. This periodic schedule limits the influence of artificial divergence but does not eliminate proxy failures.

\section{Experiments}

\begin{table*}[t]
\centering

\resizebox{\textwidth}{!}{
\begin{tabular}{l|cc|cc|cc|c|c}
\toprule
\multirow{2}{*}{Method} & \multicolumn{2}{c|}{AMC23} &  \multicolumn{2}{c|}{AIME25} & \multicolumn{2}{c|}{AIME24} &  MATH & \multirow{2}{*}{Average} \\
\cmidrule(lr){2-3} \cmidrule(lr){4-5} \cmidrule(lr){6-7} \cmidrule(lr){8-8}
& pass@1 & pass@16 & pass@1 & pass@16 & pass@1 & pass@16 & pass@1 & \\
\midrule
Base Model &  8.1 & 51.2 & 1.3 & 10.2 & 2.9 & 16.5 & 13.9 & 14.9 \\
SFT & 39.1 & 77.3 & 4.4 & 17.2 & 6.5 & 24.6 &   65.8 & 33.6 \\
\midrule
GRPO & 61.4 & \underline{88.8} & 16.0 & 33.3 & 17.9 & 39.9 &  82.9 & 48.6 \\
LTE \citep{tang2026LTE} &  56.7 & 85.0 & 14.2 & \textbf{46.7} & \underline{18.3} & 30.0 &  74.6 & 46.5 \\
TTRL \citep{zuo2025ttrl} &  42.5 & 75.2 & 6.8 & 28.6 & 12.1 & 23.2 &  75.4 & 37.7 \\
EVOL-RL  \citep{zhou2025evolrl} & 51.4 & 80.3 & 17.5 & 39.9 & \textbf{20.5} & \textbf{47.6} &   80.0 & 48.2 \\
\midrule
Parallel-R1 (Seen) &  62.0 & \underline{88.8} & 18.1 & 36.3 & 17.1 & 35.1 & 82.0 & 48.5 \\
Parallel-R1 (Unseen) & \underline{63.6} & 88.2 & \underline{18.3} & 38.5 & 16.5 & 37.1 &  83.2 & \underline{49.3} \\
Parallel-R1 (w/ Jaccard) & 62.0 &	85.2 & 18.1 & 38.1 & 17.1 &	31.8 & 82.6 & 47.8 \\
\midrule
DPT (Seen) & 62.0 & 88.7 & 17.7 & 38.4 & \underline{18.3} & 36.6 &   \underline{83.5} & \underline{49.3} \\
\textbf{DPT (Unseen)} & \textbf{64.4} & \textbf{89.9} &  \textbf{18.5} & \underline{40.2} & 18.1 & \underline{43.3} &  \textbf{84.5} & \textbf{51.3} \\
\bottomrule
\end{tabular}
}
\caption{Performance comparison on mathematical reasoning benchmarks, where the best results shown in bold and second-best results underlined.}
\label{tab:main_results}
\end{table*}

In this section, we evaluate DPT on multiple mathematical reasoning benchmarks.We conduct experiments on coding tasks and different LLM backbones to demonstrate the effectiveness of DPT and the necessity of path diversity in parallel thinking. 

\paragraph{Datasets.} For reinforcement learning, we train on the DAPO dataset~\citep{yu2025dapo} using GRPO. We evaluate mainly on four challenging mathematical reasoning benchmarks: AIME24, AIME25, AMC23~\citep{AMC23} and MATH~\citep{hendrycks2021math}. We report pass@1 and pass@16 accuracy, where pass@k measures the fraction of problems solved correctly in at least one of k sampled solutions.  We also conduct experiments on HumanEval~\citep{liu2023your}, ARC-c~\citep{clark2018think}, MMLU-Pro~\citep{wang2024mmlu} to evluate the generalization of DPT on coding and knowledge reasoning tasks.

\paragraph{Baselines.} We compare against nine baselines: Direct, Cold-start SFT, GRPO, LTE~\citep{tang2026LTE}, TTRL~\citep{zuo2025ttrl}, EVOL-RL~\citep{zhou2025evolrl}, Parallel-R1 (Seen/Unseen) and Parallel-R1 (w/ Jaccard). Unseen mode uses structured attention and multiverse position encodings to prevent cross-path information leakage. LTE, TTRL and EVOL-RL results are reported from their original papers. 

\paragraph{Implementation Details.} The main experiments are conducted using 8 A100 GPUs with Qwen-3-4B-Base~\citep{yang2025qwen3technicalreport}.  Our codebase is adapted from VERL~\citep{Sheng_2025verl}. For cold-start initialization, we construct parallel-thinking examples from GSM8K by prompting  DeepSeek-R1-0528-Qwen-3-8B and filtering outputs for valid tag structure. For RL, we train on DAPO~\citep{yu2025dapo} using GRPO. The main configuration uses $K=2$, $\lambda_f=0.2$, $\lambda_d=0.1$, and $n=1$.
Details can be found in Appendix \ref{sec: implementation details}.


\subsection{Experimental Results}

\paragraph{Overall Performance.} Table~\ref{tab:main_results} shows that DPT (Unseen) achieves the best overall performance among the evaluated methods, outperforming Parallel-R1 (Unseen) across every reported benchmark and metric. The gains span both single-sample accuracy and multi-sample coverage, with a particularly strong improvement on challenging AIME problems. For instance, DPT improves AIME24 pass@16 by 6.2 percentage points over Parallel-R1 (Unseen), highlighting its ability to uncover correct solutions through broader exploration.
DPT (Unseen) also outperforms DPT (Seen) on average, suggesting that maintaining independent paths complements diversity-driven exploration. We use DPT to denote the Unseen configuration in the following experiments.

\paragraph{Reasoning Beyond Mathematics.}
\label{sec:additional_domains}
We extend the evaluation to code generation on HumanEval, science and commonsense reasoning on ARC-c, and knowledge-intensive reasoning on MMLU-Pro. As shown in Table~\ref{tab:additional_domains}, DPT outperforms both GRPO and Parallel-R1 on all three benchmarks. The advantages are most pronounced on code generation and knowledge-intensive reasoning, while DPT also maintains the lead on ARC-c. These consistent gains broaden the applicability of our approach. Encouraging diverse internal paths benefits reasoning across tasks with different solution structures and correctness signals. In particular, the improvement over Parallel-R1 highlights the additional value of DPT within the parallel-thinking framework beyond mathematics.

\begin{table}[t]
\centering
\small
\setlength{\tabcolsep}{4pt}

\begin{tabular}{@{}lrrr@{}}
\toprule
Method & HumanEval & ARC-c & MMLU-Pro \\
\midrule
GRPO & 60.98 & 64.68 & 43.40 \\
Parallel-R1 & 67.12 & 64.88 & 47.76 \\
DPT & \textbf{69.51} & \textbf{65.27} & \textbf{49.40} \\
\bottomrule
\end{tabular}
\caption{Reported scores (\%) on three additional on coding and knowledge reasoning tasks.}
\label{tab:additional_domains}
\end{table}



\begin{table}[t]
\centering

\vspace{0.5em}
\small
\setlength{\tabcolsep}{3pt}
\begin{tabular}{@{}lcccc@{}}
\toprule
Method & AIME24 & AIME25 & AMC23 & MATH \\
\midrule
Parallel-R1 & 0.581 & 0.542 & 0.529 & 0.567 \\
\textbf{DPT} & \textbf{0.923} & \textbf{0.927} & \textbf{0.911} & \textbf{0.886} \\
$\Delta$ & $\uparrow$ 0.342 & $\uparrow$ 0.385 & $\uparrow$ 0.382 & $\uparrow$ 0.319 \\
\bottomrule
\end{tabular}
\caption{Lexical path diversity measured by Jaccard distance. Higher values indicate lower lexical overlap; $\Delta$ denotes the increase over Parallel-R1.}
\label{tab:diversity}
\end{table}

\begin{figure}[t]
\centering
\includegraphics[width=0.9\linewidth]{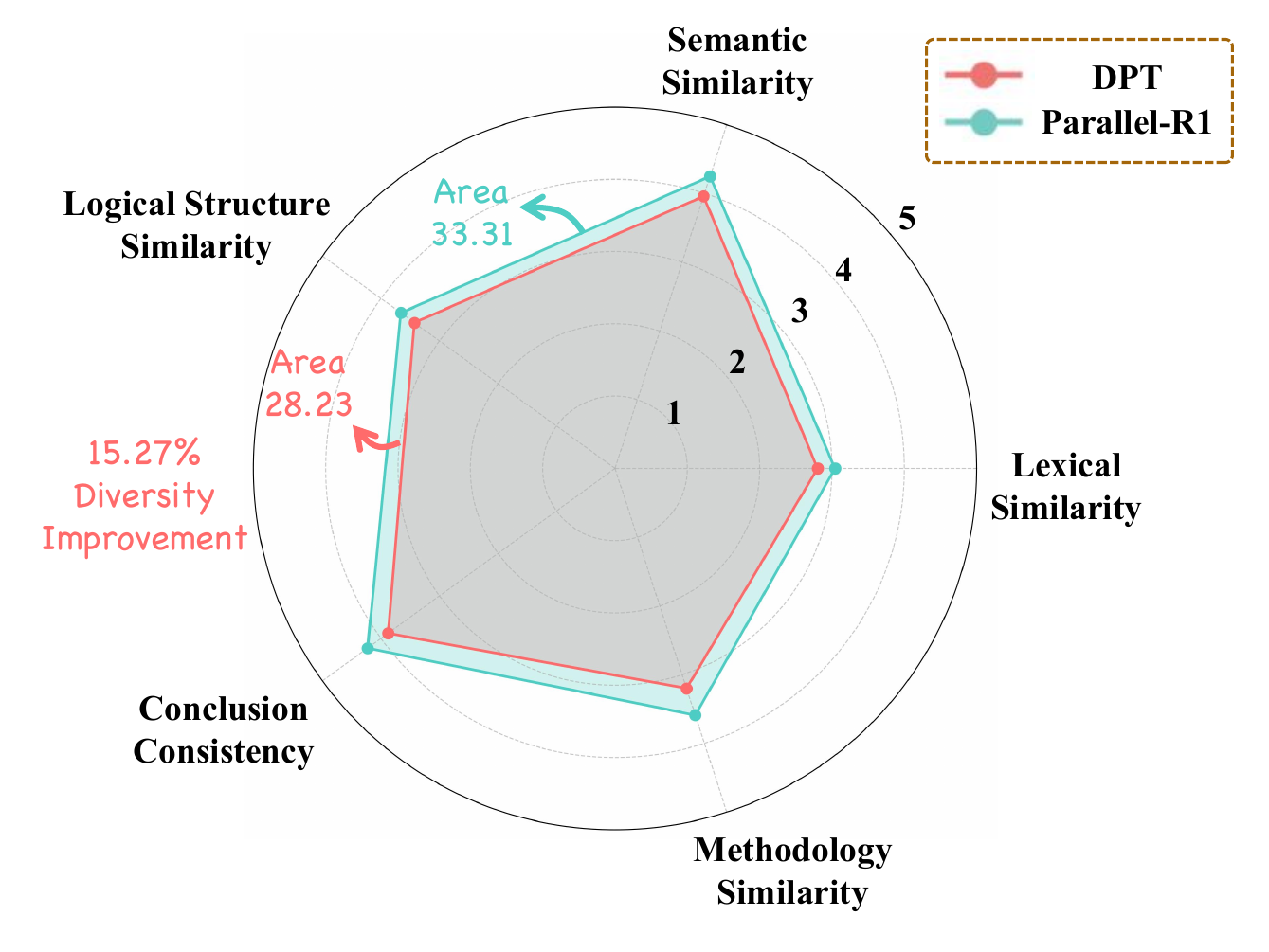}
\caption{\textbf{Multi-dimensional diversity evaluation.} LLM-as-judge comparing DPT and Parallel-R1 across five dimensions. Lower score denotes more diversity.}
\label{fig:llm_judge}
\end{figure}

\paragraph{Path Diversity.} Beyond accuracy, DPT produces substantially less overlapping reasoning paths across all four mathematical benchmarks (Table~\ref{tab:diversity}). Although our reward acts only on path prefixes, the increase in whole-path Jaccard distance shows that early divergence is accompanied by broader differences in the generated continuations. This local intervention thus encourages broader textual exploration while leaving the remainder of each path free to develop.
The complementary LLM-as-judge assessment in Figure~\ref{fig:llm_judge} also favors DPT across solution approach, reasoning structure, mathematical techniques, explanation style, and strategy diversity. Together, these observations support our goal of making parallel branches more distinct and useful for exploration. Appendix~\ref{sec:lexical_semantic_diversity} provides further details on the lexical and judge-based evaluations.

\begin{table*}[t]
\centering
\small

\begin{tabular}{lrrrrrr}
\toprule
& \multicolumn{2}{c}{AIME25} & \multicolumn{2}{c}{AIME24} & \multicolumn{2}{c}{AMC23} \\
\cmidrule(lr){2-3}\cmidrule(lr){4-5}\cmidrule(lr){6-7}
Setting & pass@1 & pass@16 & pass@1 & pass@16 & pass@1 & pass@16 \\
\midrule
DPT ($K=2$) & \textbf{18.5} & 40.2 & \textbf{18.1} & \textbf{40.2} & \textbf{64.4} & 89.9 \\
DPT ($K=4$) & 17.5 & \textbf{41.2} & 17.5 & 38.2 & 61.9 & \textbf{90.4} \\
\bottomrule
\end{tabular}
\caption{Results (\%) with two or four internal reasoning paths. $K$ counts paths within a trajectory. We report pass@1 and pass@16 accuracy on three mathematical reasoning benchmarks.}
\label{tab:additional_paths}
\end{table*}

\subsection{Further Analysis}

\paragraph{Scaling the Number of Internal Paths.}
\label{sec:additional_paths}

We further explore scaling within a single trajectory by increasing the number of internal paths from $K=2$ to $K=4$ under the same 4B backbone and training setup. Our normalized pairwise diversity reward extends directly to more paths without changing its definition. Table~\ref{tab:additional_paths} reveals a clear division of strengths: $K=2$ delivers the best pass@1 accuracy on all three benchmarks, while $K=4$ achieves higher pass@16 on AIME25 and AMC23 and remains competitive on AIME24. This pattern shows that additional paths strengthen exploration and solution coverage, even when a smaller path set is preferable for the most likely individual answer. 
\paragraph{Test-time Scaling.} DPT maintains the strongest performance at every sampling budget in Table~\ref{tab:passk}, combining higher single-sample accuracy with sustained gains as more complete solutions are sampled. Parallel-R1's initial advantage over GRPO diminishes at larger budgets, whereas DPT continues to outperform both methods. At pass@256, DPT exceeds Parallel-R1 by 6.1 percentage points. This pattern is consistent with diversity helping parallel thinking retain broader solution coverage as sampling increases. We further discuss this interpretation in Appendix~\ref{sec:scaling_interpretation} and report inference-cost measurements in Appendix~\ref{sec:additional_cost}. Together with the paths scaling results, it illustrates how DPT supports broader exploration both within a trajectory and across sampled solutions.

\begin{table}[t]
\centering

\small
\setlength{\tabcolsep}{3pt}
\begin{tabular}{@{}lrrrrr@{}}
\toprule
Method & \multicolumn{5}{c}{pass@$k$ (\%)} \\
\cmidrule(lr){2-6}
& $1$ & $16$ & $64$ & $128$ & $256$ \\
\midrule
GRPO & 16.0 & 33.3 & 48.3 & 52.4 & 56.0 \\
Parallel-R1 & 18.3 & 38.5 & 47.9 & 49.3 & 50.9 \\
\textbf{DPT} & \textbf{18.5} & \textbf{40.2} & \textbf{49.2} & \textbf{53.5} & \textbf{57.0} \\
\bottomrule
\end{tabular}
\caption{AIME25 pass@$k$ accuracy (\%) with increasing numbers of sampled complete solutions.}
\label{tab:passk}
\end{table}

\begin{figure}[t]
\centering
\includegraphics[width=\linewidth]{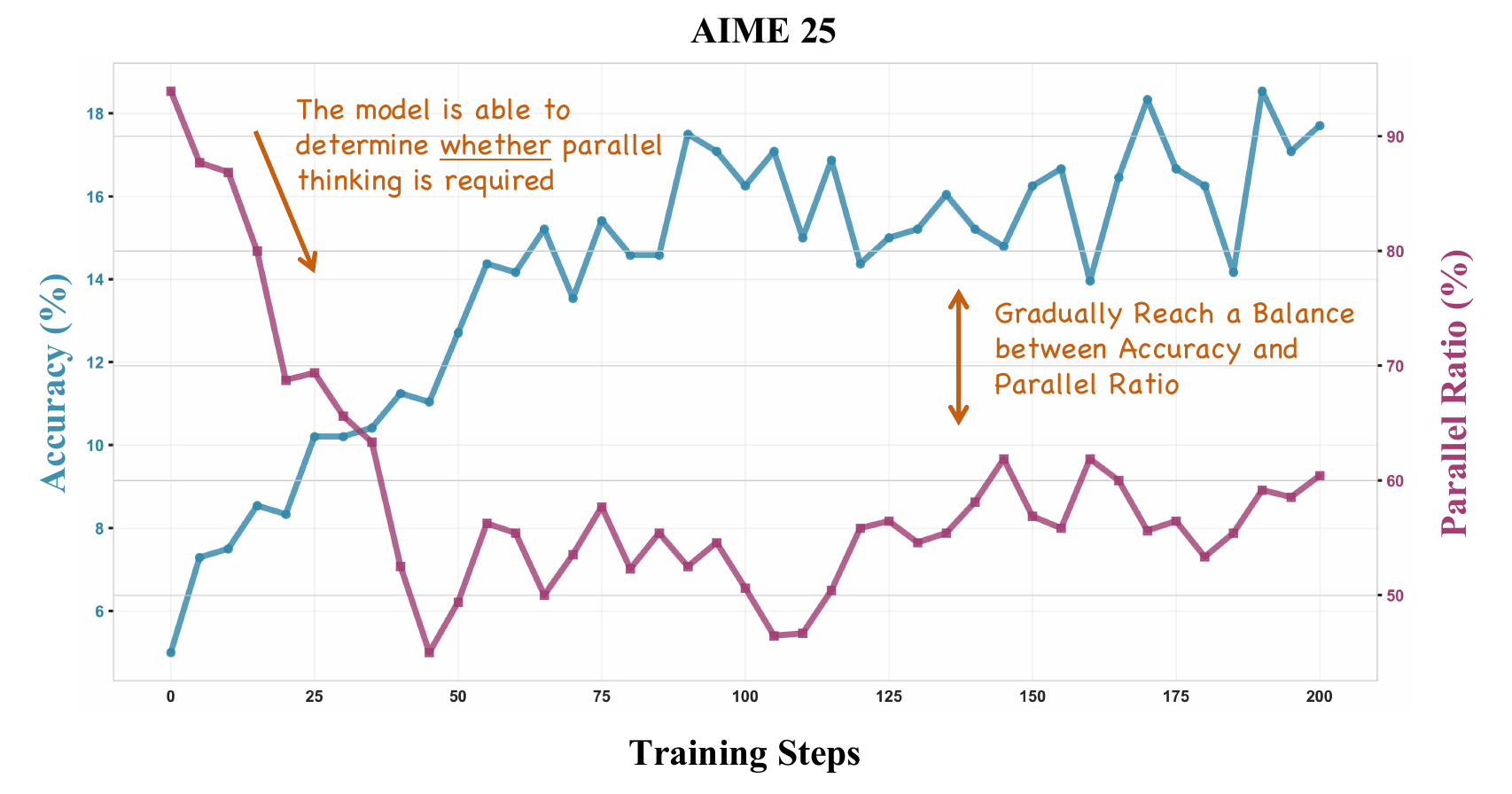}
\caption{Training dynamics of accuracy and parallel ratio, where the parallel ratio (purple) decreases from near 100\% to $\sim$65\% as the model learns selective parallel thinking, while accuracy (blue) continues improving.}
\label{fig:training_dynamics}
\end{figure}

\paragraph{Adaptive Parallel Thinking Usage.} Figure~\ref{fig:training_dynamics} shows that the model shifts from near-universal parallel thinking after cold-start initialization to more selective use during RL training, while accuracy continues to improve. The trained model increasingly combines selective branching with effective solution generation. This behavior complements our diversity objective: DPT encourages distinct paths when the model branches, while the training process allows it to adapt how frequently it invokes parallel thinking. We provide the ablation of prefix length and reward design in Appendix~\ref{sec: full_ablation}.

\section{Conclusion}
In this paper, we first discover the problem of path mode collapse in parallel thinking, where multiple parallel reasoning paths degenerate into similar ones, contradicting the core motivation of parallel thinking. Therefore, we propose a simple but effective RL approach, DPT, to address path mode collapse in parallel thinking for LLMs. Specifically, DPT enforces path diversity through a token-level diversity metric that detects early reasoning divergence, combined with a two-stage training curriculum that balances correctness and exploration. Extensive experiments demonstrate that DPT consistently outperforms the best-performing baseline on mathematical reasoning, coding and knowledge reasoning benchmarks.
The results establish that path diversity is a key component for effective parallel thinking.
By addressing path mode collapse, we take a step toward realizing the full potential of parallel thinking, unlocking the test-time scaling.

\section*{Limitations}
Our study focuses on text-based reasoning tasks with verifiable answers, including mathematics, coding, and knowledge reasoning. Extending DPT to multimodal inputs and interactive reasoning would further broaden its application scope. We evaluate a small set of internal path counts and keep the path count fixed within each configuration. An adaptive allocation policy that selects the number of paths according to problem difficulty and inference budget could provide greater flexibility. We leave these directions for future work.

\section*{Ethical Considerations}
Within the evaluated research setting, we do not identify substantial additional ethical, societal, or environmental risks beyond those generally associated with LLM training and inference.

\bibliography{example_paper}

\appendix

\section{Implementation Details}
\label{sec: implementation details}
The main experiments are conducted using 8 A100 GPUs with Qwen-3-4B-Base~\citep{yang2025qwen3technicalreport}, following concurrent parallel-thinking research~\citep{zheng2025parallelr1,wu2025nativeparallelreasoner}. The objective-level discussion in Section~\ref{sec:motivation} motivates the reward design without establishing architecture-invariant behavior or inevitable collapse. Our codebase is adapted from VERL~\citep{Sheng_2025verl}; we follow its training recipe and separately ablate prefix length and comparison method. For cold-start initialization, we construct parallel-thinking examples from GSM8K by prompting a capable model and filtering outputs for valid tag structure. We perform supervised fine-tuning with batch size 128, learning rate 1e-5, weight decay
of 0.01, and warm-up ratio 0.1 with a cosine learning-rate schedule. For RL, we train on DAPO~\citep{yu2025dapo} using GRPO with batch size 256, learning rate 1e-6, and 300 gradient update steps. The main configuration uses $K=2$, $\lambda_f=0.2$, $\lambda_d=0.1$, and Prefix-1 exact matching ($n=1$). Diversity and format rewards are applied only for the last two steps of each ten-step cycle; the remaining eight use correctness alone. This periodic RL schedule follows cold-start SFT.

\section{Evaluation Benchmarks}
\label{sec: Evaluation Benchmarks}
\begin{figure}[t]
    \centering
    \includegraphics[width=0.5\textwidth]{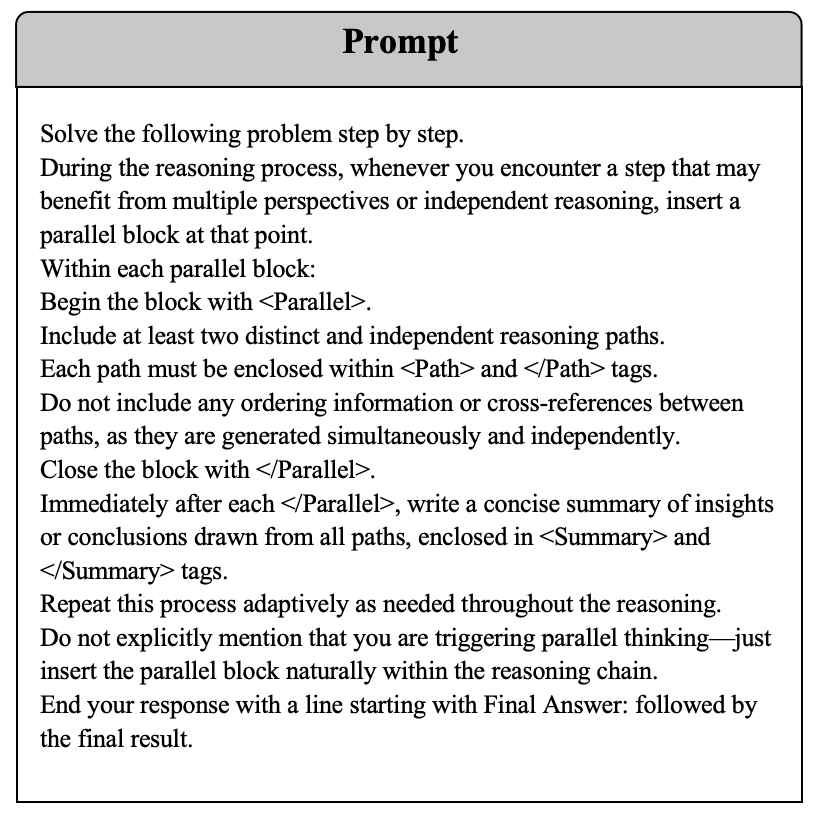}
    \caption{Prompt Construction.}
    \label{fig: prompt}
\end{figure}

We evaluate on four challenging mathematical reasoning benchmarks. AIME~\citep{AIME} (American Invitational Mathematics Examination) tests advanced problem-solving with AIME24 and AIME25 versions. AMC23~\citep{AMC23} (American Mathematics Competition) evaluates mathematical competition skills. MATH~\citep{hendrycks2021math} contains high-difficulty problems across multiple mathematical domains. Following prior work, we report pass@1 and pass@16 accuracy, where pass@k measures the fraction of problems solved correctly in at least one of k sampled solutions. For AIME and AMC benchmarks, we sample 16 responses per question at temperature 1.0 and report mean accuracy. For MATH, we evaluate on a 300-problem subset and report pass@1 with single sampling.
For DPT and other parallel thinking baselines, we construct specific instruction prompt for all problems in the training and testing datasets to guide the model to conduct parallel thinking, as shown in \cref{fig: prompt}.

\section{Baselines}
\label{sec: baselines}
In this section, we introduce three RL baselines adopted in our experiment: LTE~\citep{tang2026LTE}, TTRL~\citep{zuo2025ttrl} and EVOL-RL~\citep{zhou2025evolrl}.

\noindent{\textbf{LTE:}} LTE (Learning to Reason from Trial and Error) addresses the exploration stagnation issue in reinforcement learning with verifiable rewards (RLVR)~\citep{wang2025reinforcementlearningreasoninglarge,wen2025reinforcementlearningverifiablerewards}, where models are often limited by their initial on-policy generation capability. Unlike standard RLVR methods that rely solely on on-policy responses or off-policy approaches requiring external expert solutions, LTE leverages models' own previously generated mistakes as learning signals, without any external supervision. By incorporating trial-and-error feedback into training, LTE improves both exploration and exploitation. LTE consistently outperforms GRPO  on multiple mathematical and even surpasses methods using external gold solutions under aligned settings, demonstrating its effectiveness in enhancing reasoning performance.

\noindent{\textbf{TTRL:}} TTRL (Test-Time Reinforcement Learning) studies reinforcement learning for LLM reasoning tasks in the absence of explicit supervision, where ground-truth labels are unavailable and reward estimation becomes the central challenge. Instead of relying on oracle rewards, TTRL exploits implicit rewards derived from common test-time scaling strategies~\citep{muennighoff2025s1,zhang2025survey}, such as majority voting~\citep{gudatiny,kamen2025majority}, to guide RL training on unlabeled data. By leveraging the inductive priors of pre-trained models, TTRL enables self-improvement during inference without additional annotations. Extensive experiments show that TTRL consistently improves performance across tasks and model scales, and achieves substantial gains on challenging benchmarks such as AIME 2024. Notably, despite being supervised only by majority-based rewards, TTRL surpasses the initial model's maj@n upper bound and approaches the performance of models trained with ground-truth supervision, demonstrating the effectiveness of test-time RL with implicit rewards.

\noindent{\textbf{EVOL-RL:}} EVOL-RL targets label-free self-improvement for LLMs trained with reinforcement learning from verifiable rewards, where no ground-truth labels or external judges are available at deployment time. Unlike prior self-improvement methods that rely on self-confirmation signals such as confidence or majority consistency~\citep{weng2023largelanguagemodelsbetter,zhang2025incentivizingllmsselfverifyanswers, becker2024cyclesthoughtmeasuringllm}, EVOL-RL adopts an evolution-inspired training paradigm. It stabilizes learning by anchoring rewards to the majority-voted solution, while simultaneously encouraging exploration through a novelty-aware reward that promotes reasoning diversity among sampled responses. This balance between majority-based selection and novelty-driven variation prevents over-confident collapse and sustains complex reasoning behaviors. Experimental results demonstrate that EVOL-RL consistently outperforms majority-only baselines on label-free reasoning benchmarks and improves both in-domain performance and out-of-domain generalization across diverse tasks.

\section{Ablation on Diversity Metric Design}
\label{sec: full_ablation}

\paragraph{Exact and Similarity Settings.} Both settings compare the first $n$ content tokens after the \texttt{<Path>} tag. Let $u_i=\text{prefix}(\tau_i,n)$ and $u_j=\text{prefix}(\tau_j,n)$ denote the two prefixes. The \textbf{exact} setting uses the binary score:
\begin{equation}
d_{\mathrm{exact}}(\tau_i,\tau_j)=\mathbb{I}[u_i\neq u_j].
\end{equation}
It assigns a diversity score of 1 whenever at least one corresponding token differs, and 0 only when the entire prefixes are identical. The \textbf{similarity} setting instead assigns the proportion of corresponding token positions that differ:
\begin{equation}
d_{\mathrm{sim}}(\tau_i,\tau_j)=\frac{1}{n}\sum_{t=1}^{n}\mathbb{I}[(u_i)_t\neq(u_j)_t].
\end{equation}
For example, if two three-token prefixes differ at exactly one position, exact gives a score of 1, whereas similarity gives $1/3$. Thus, exact rewards any prefix divergence, while similarity provides graded credit according to its extent. The two variants coincide when $n=1$. For either variant, the trajectory-level diversity reward averages these scores over all pairs of internal paths, as in the main text.

Table~\ref{tab:ablation} compares prefix lengths and the exact and similarity settings across all four mathematical reasoning benchmarks. Every tested prefix-diversity setting improves average accuracy over Parallel-R1 without a diversity reward, supporting the effectiveness of early path divergence across different reward designs. Prefix-1 exact achieves the highest average accuracy and improves every reported accuracy metric over the baseline. We therefore select $n=1$ with exact matching for our main configuration based on its strongest aggregate performance across the full evaluation suite.

\begin{table*}[t]
\centering
\small
\setlength{\tabcolsep}{3pt}
\begin{tabular}{@{}lrrrrrrrr@{}}
\toprule
\multirow{2}{*}{Metric} & \multicolumn{2}{c}{AIME25} & \multicolumn{2}{c}{AIME24} & \multicolumn{2}{c}{AMC23} & MATH & \multirow{2}{*}{Average} \\
\cmidrule(lr){2-3} \cmidrule(lr){4-5} \cmidrule(lr){6-7} \cmidrule(lr){8-8}
& pass@1 & pass@16 & pass@1 & pass@16 & pass@1 & pass@16 & pass@1 & \\
\midrule
No diversity reward & 18.33 & 38.51 & 16.45 & 37.11 & 63.59 & 88.20 & 83.22 & 49.34 \\
\midrule
Prefix-1 exact & 18.54 & 40.18 & \textbf{18.12} & \textbf{43.34} & 64.37 & 89.88 & 84.49 & \textbf{51.27} \\
Prefix-2 exact & 17.91 & 42.56 & 16.66 & 35.88 & 62.81 & 88.50 & \textbf{85.12} & 49.92 \\
Prefix-3 exact & 17.08 & 40.92 & 17.70 & 33.72 & 63.75 & 89.10 & 84.49 & 49.54 \\
Prefix-4 exact & \textbf{20.83} & 42.88 & 17.91 & 36.77 & \textbf{64.53} & \textbf{91.26} & 82.59 & 50.97 \\
\midrule
Prefix-2 similarity & 19.37 & 42.50 & 17.70 & 36.86 & 63.12 & 89.42 & 83.22 & 50.31 \\
Prefix-3 similarity & 17.70 & 39.82 & 17.50 & 39.51 & 63.12 & 90.25 & 83.22 & 50.16 \\
Prefix-4 similarity & \textbf{20.83} & \textbf{46.27} & 17.29 & 40.16 & 63.58 & 87.39 & 82.75 & 51.18 \\
\bottomrule
\end{tabular}
\caption{Ablation study on diversity metric design across four mathematical reasoning benchmarks. Scores are percentages; Average is the arithmetic mean of the seven reported accuracy metrics. The best results in each column are bold.}
\label{tab:ablation}
\end{table*}

Longer prefixes offer advantages on individual benchmarks: Prefix-4 similarity achieves the strongest AIME25 pass@16 result, Prefix-4 exact leads on AMC23, and Prefix-2 exact performs best on MATH. Prefix-1 exact combines the strongest AIME24 results with improvements across the remaining benchmarks, yielding the best overall average of 51.27\%. Its selection thus reflects performance across the full suite rather than optimizing for a single benchmark. Table~\ref{tab:ablation_parallel_ratio} additionally shows that the configuration with the highest parallel-thinking usage on AIME25 and AIME24 is Prefix-2 exact. This distinction reinforces that effective path exploration, rather than branching frequency alone, is important for overall reasoning performance.

\begin{table}[t]
\centering
\small
\setlength{\tabcolsep}{3pt}
\begin{tabular}{@{}lrrrr@{}}
\toprule
Metric & AIME25 & AIME24 & AMC23 & MATH \\
\midrule
No diversity reward & 54.58 & 52.08 & 77.81 & 94.62 \\
\midrule
Prefix-1 exact & 60.42 & 61.67 & \textbf{82.50} & \textbf{95.89} \\
Prefix-2 exact & \textbf{75.00} & \textbf{77.71} & 81.80 & \textbf{95.89} \\
Prefix-3 exact & 48.75 & 49.17 & 73.75 & 92.72 \\
Prefix-4 exact & 51.88 & 51.04 & 77.34 & 89.87 \\
\midrule
Prefix-2 similarity & 42.08 & 44.79 & 71.88 & 92.72 \\
Prefix-3 similarity & 38.12 & 41.67 & 70.94 & 87.66 \\
Prefix-4 similarity & 52.92 & 55.21 & 77.34 & 93.35 \\
\bottomrule
\end{tabular}
\caption{Parallel ratio (\%) for each diversity metric setting, measuring how often the model triggers parallel thinking during the reasoning process.}
\label{tab:ablation_parallel_ratio}
\end{table}

\section{Interpreting Test-Time Scaling}
\label{sec:scaling_interpretation}

The pass@$k$ curves in Table~\ref{tab:passk} highlight DPT's ability to convert path diversity into stronger test-time scaling. DPT is competitive at pass@1 and becomes increasingly advantageous as more complete solutions are sampled, remaining ahead of both baselines at every larger budget. From $k=64$ to $k=256$, DPT rises from 49.2 to 57.0, compared with 47.9 to 50.9 for Parallel-R1, demonstrating broader coverage of correct solution modes.

The scaling pattern is naturally explained by DPT's reduced redundancy: its internal paths cover more distinct solution directions, so additional complete samples continue to contribute useful coverage rather than revisiting the same modes. The agreement between the path-diversity measurements and the pass@$k$ curves supports diversity as a central mechanism behind DPT's gains in both within-response exploration and test-time scaling.

\section{Inference Cost and Response Length}
\label{sec:additional_cost}

The diversity reward is used only during training. DPT adds no inference-time judge, reward module, or additional model call beyond the native parallel-thinking generation process. We analyze 1,916 validation trajectories at checkpoint 175 after removing trailing padding. Table~\ref{tab:additional_cost} reports the measured token counts and latency together with the reported FLOP estimates.

\begin{table*}[t]
\centering
\small
\begin{tabular}{lrrrr}
\toprule
Method & Tokens/problem & Latency (s) & TFLOPs/problem & Tokens/solved problem \\
\midrule
GRPO (sequential) & 1,783 & 0.778 & 17.25 & \textbf{3,544} \\
Parallel-R1 & 1,666 & 0.710 & 15.64 & 4,004 \\
DPT & \textbf{1,584} & \textbf{0.678} & \textbf{14.99} & 3,785 \\
\bottomrule
\end{tabular}
\caption{Inference cost in the checkpoint-175 validation analysis. FLOPs are reported in trillions per problem. The last column normalizes token consumption by solved problems.}
\label{tab:additional_cost}
\end{table*}

Relative to Parallel-R1, DPT reduces tokens per problem by 4.9\%, latency by 4.5\%, and FLOPs by 4.2\%, while also reducing tokens per solved problem by 5.5\%. Thus, DPT combines stronger reasoning performance with a leaner inference profile within the parallel-thinking setting. The comparison with sequential GRPO further shows that DPT achieves its gains with substantially fewer raw tokens, highlighting the practical value of selective, diverse internal exploration.

Response length provides an intuitive explanation for this efficiency profile. DPT produces the shortest median responses and reaches the 3,000-token cap least often (Table~\ref{tab:additional_lengths}), indicating that its diversity reward helps the model explore alternatives without unnecessarily prolonging each branch. The combination of concurrent generation and concise paths gives DPT a favorable latency and token profile while preserving the benefits of parallel exploration.

\begin{table}[t]
\centering
\small
\begin{tabular}{lrr}
\toprule
Method & Median tokens & At token cap (\%) \\
\midrule
GRPO & 1,715 & 24.0 \\
Parallel-R1 & 1,491 & 13.0 \\
DPT & 1,401 & 10.2 \\
\bottomrule
\end{tabular}
\caption{Response lengths and frequency of reaching the 3,000-token cap in the validation cost analysis.}
\label{tab:additional_lengths}
\end{table}

Overall, the measurements show that DPT's training-time diversity objective carries through to an efficient inference profile: the model generates shorter responses, uses fewer tokens, and maintains the strongest reasoning results in the evaluated parallel setting. These properties make DPT attractive for deployments where both solution coverage and response efficiency matter.

\section{Prefix Divergence as a Training Proxy}
\label{sec:prefix_proxy}

Prefix diversity provides a low-cost and effective control point for shaping exploration. Because early tokens establish the conditional context for the remainder of an autoregressive path, encouraging distinct prefixes gives parallel branches different starting points from which to develop. The resulting signal is simple to compute, easy to interpret, and directly aligned with the goal of avoiding duplicated internal reasoning.

The practical rationale is local symmetry breaking. For paths sampled from the same preceding context, changing an initial token changes the conditional distribution of subsequent tokens, creating room for branches to develop different continuations instead of immediately reproducing the same trajectory. This gives DPT a targeted exploration signal while leaving most of each path unconstrained to follow the task reward.

The prefix length $n$ controls the extent of the intervention. The default $n=1$ encourages branches to make different initial decisions while leaving almost all subsequent tokens governed by the task objective. Larger prefixes extend the exploration window, and the similarity variant provides graded credit according to how many positions differ. The ablation shows that these choices offer complementary operating points for balancing immediate accuracy and broader solution coverage.

The ablation in Appendix~\ref{sec: full_ablation} illustrates these design choices empirically. Prefix-1 exact achieves the highest average accuracy across the full evaluation suite and is therefore selected as the default configuration. The strengths of other settings on individual benchmarks further motivate treating prefix length and comparison granularity as tunable components.
Correctness remains the primary task signal throughout training. Diversity and format rewards are applied for the last two steps of each ten-step RL cycle, while the remaining eight steps use correctness alone. This curriculum lets DPT introduce exploration periodically while continuously anchoring learning to verifiable task success. The same design also leaves room for future extensions such as adaptive prefix selection and strategy-aware objectives.

\section{Lexical Overlap and Semantic Evaluation}
\label{sec:lexical_semantic_diversity}

Whole-path Jaccard distance is useful as a diagnostic of the diversity induced by training. As a training reward, however, it couples the signal to path length and lexical choices. The controlled comparison in Table~\ref{tab:main_results} shows that the prefix formulation yields a higher average score than both the whole-path Jaccard variant and Parallel-R1 without diversity regularization, supporting the value of applying the signal at the point where branches first separate.
As a post-hoc diagnostic, Jaccard distance reveals how the local prefix reward propagates through complete responses. Table~\ref{tab:diversity} shows substantially larger whole-path distances for DPT across all four benchmarks, indicating that early branching decisions lead to broader textual exploration throughout the generated solutions.

The 1,156 blocks used in the motivation analysis originate from validation rollouts of the reproduced Parallel-R1 model at training step 200. They cover AIME24, AIME25, AMC23, and MATH. Each included block is a generated \texttt{<Parallel>} span containing two parsed \texttt{<Path>} spans; malformed or incomplete parallel spans are excluded. 

GPT-5-mini compares each pair on solution approach, reasoning structure, mathematical technique, expression style, and overall strategy diversity. It also provides an explanation for each judgment, all of which are manually reviewed to ensure reasonableness and fairness. DPT is favored across these dimensions, complementing the lexical results with evidence that its paths differ in their reasoning organization as well as their surface form. Together, the two views show that the prefix reward produces diverse responses at both the token and solution-structure levels.




\section{Case Studies}
\begin{figure}[t]
    \centering
    \includegraphics[width=0.5\textwidth]{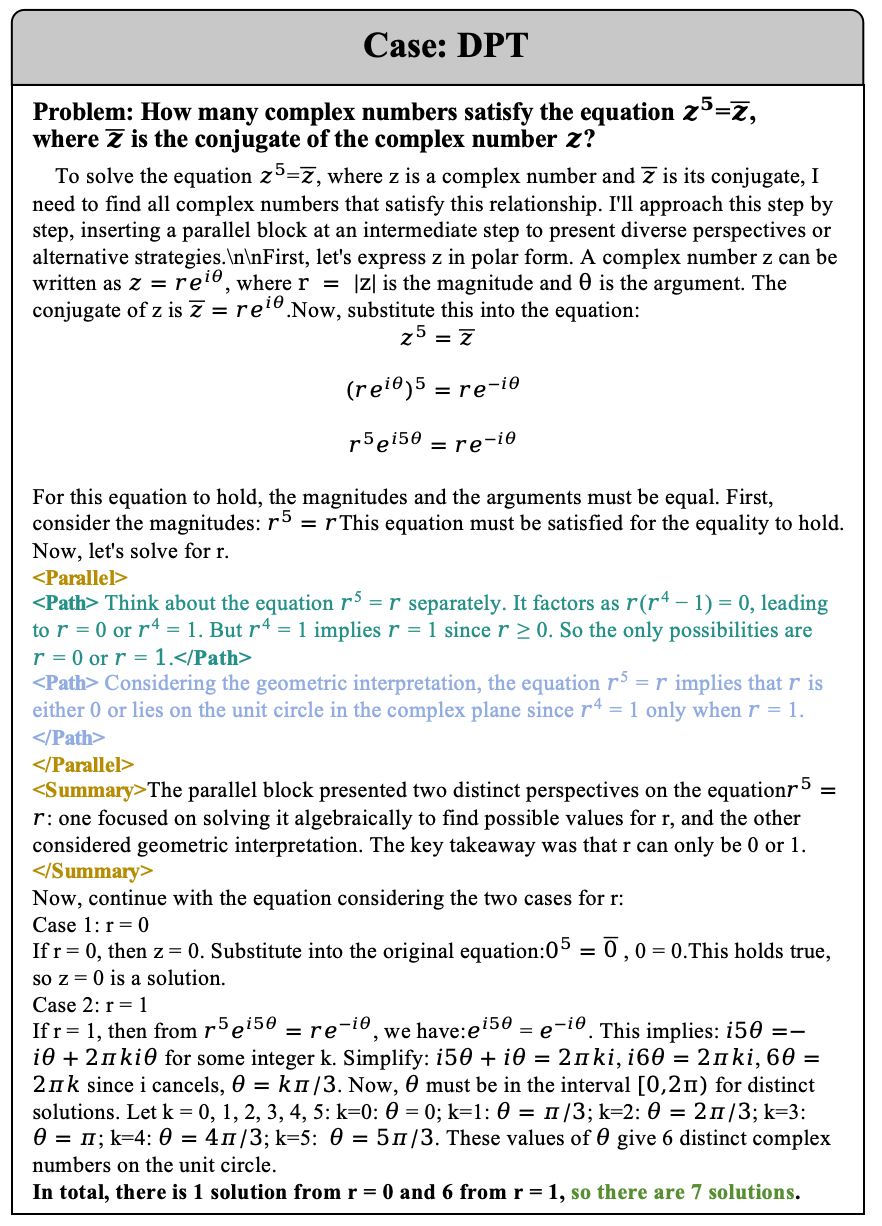}
    \caption{Case Study of DPT}
    \label{fig: case_dpt}
\end{figure}

\begin{figure}[t]
    \centering
    \includegraphics[width=0.5\textwidth]{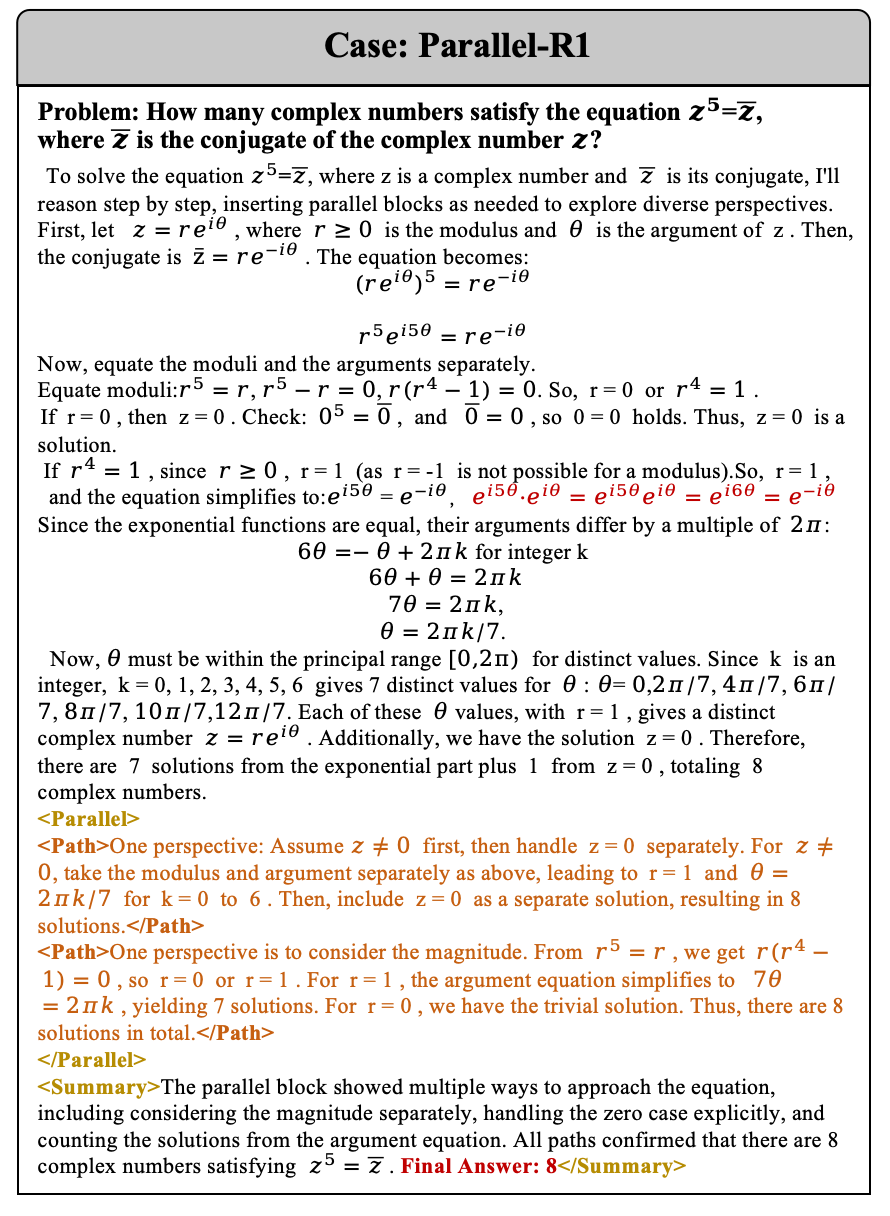}
    \caption{Case Study of Parallel-R1}
    \label{fig: case_r1}
\end{figure}

In this section, we conduct a detailed analysis of our DPT and Parallel-R1 on a specific test case. For this complex-number computation problem, our DPT performs parallel reasoning by approaching the solution for \( r \) from the perspective of \( r^5 = r \) as shown in \cref{fig: case_dpt}. In particular, the two reasoning paths solve the problem from algebraic and geometric viewpoints, respectively. Although the unit-circle remark in the second path is imprecise, it still helps explore potential possibilities in the solution space. In contrast, in Parallel-R1, the model first makes an incorrect calculation (highlighted in red in \cref{fig: case_r1}). During parallel reasoning, the two paths rely on the same core reasoning strategy: separating the modulus and the argument, identifying \( r = 0 \) or \( r = 1 \), deriving \( \theta = 2\pi k/7 (k = 0, \ldots, 6)\) for \( r = 1 \), and including the trivial solution \( z = 0 \), which yields eight solutions. The differences between the two paths are limited to minor presentation details rather than substantive semantic divergence. As a result, parallel reasoning fails to explore a more diverse solution space and does not correct the earlier error.

\end{document}